# Multi-Scale Deep Learning for Colon Histopathology: A Hybrid Graph-Transformer Approach

Sadra Saremi *, Amirhossein Ahmadkhan Kordbacheh*

Department of Physics, Iran University of Science and Technology, Tehran 16846-13114, Iran.

Abstract

Colon cancer also known as Colorectal cancer, is one of the most malignant types of cancer worldwide. Early-stage detection of colon cancer is highly crucial to prevent its deterioration. This research presents a hybrid multi-scale deep learning architecture that synergizes capsule networks, graph attention mechanisms, transformer modules, and residual learning to advance colon cancer classification on the Lung and Colon Cancer Histopathological Image Dataset (LC25000) dataset. The proposed model in this paper utilizes the HG-TNet model that introduces a hybrid architecture that joins strength points in transformers and convolutional neural networks to capture multi-scale features in histopathological images. Mainly, a transformer branch extracts global contextual bonds by partitioning the image into patches by convolution-based patch embedding and then processing these patches through a transformer encoder. Analogously, a dedicated CNN branch captures fine-grained, local details through successive Incorporation these diverse features, combined with a self-supervised rotation prediction objective, produce a robust diagnostic representation that surpasses standard architectures in performance. Results show better performance not only in accuracy or loss function but also in these algorithms by utilizing capsule networks to preserve spatial orders and realize how each element individually combines and forms whole structures.

Introduction

Cancer is known as one of the most common traits of human life and based on statistical research in the United States it is the second main reason for losing people's lives [1]. However, cancer mortality has been decreasing in recent years due to modifying life habits such as reforming diet plans (increasing fiber and decreasing fat [2]) and early detection of cancer types. In the case of cancer, early detection is crucial for further treatment and prevention of deterioration [3]. Colorectal cancer (CRC) is one of the most prevalent and dangerous cancer types the fourth death cancer type in 1998 became the first cancer death type in men and second in women in the present time so with early-stage detection of this type of cancer precisely can lead to save huge number of people [4,5]. Over the past several years, as a result of the emergence and development of deep learning technologies[6,7], huge advancements in medical image detection techniques have been achieved leading to significant improvement in the accuracy and efficiency of diagnosis results, convolutional neural network (CNN) has shown significant result in image processing tasks including classification, image segmentation, object detection and so on[8,9], that widely used in medical image processing to detection of malignant tumor in MRI, CT scans, X-rays or even skin lesion. The ability of hierarchical feature extraction of CNNs makes them learn complex patterns and characteristics of medical images and enables them to detect abnormalities with high accuracy [10,11]. This helps experts achieve reliable detection in early diagnosis, reducing both false negatives and false positives, while also streamlining clinical workflows. Furthermore, the integration of CNNs with other models such as transformers or Multi-Scale Graph Hybrid Capsule Transformers (MSGHCT) has emerged as a powerful strategy to enhance feature representation and overall performance [12-14]. It also provides a powerful framework that leverages the complementary strengths of each architecture. CNNs excel at capturing local features through convolutional operations, yet they often fall short in modeling long-range dependencies. In contrast, transformers use self-attention to effectively capture global contextual relationships, while MSGHCT extends this capability by incorporating capsule mechanisms and graph-based representations to capture multi-scale, hierarchical information [15,16]. This hybrid integration not only preserves the spatial detail essential for accurate feature extraction but also enhances global semantic understanding—resulting in improved performance in tasks such as medical image segmentation, fault diagnosis, and emotion recognition. Recent studies prove that utilizing such combined architectures achieves higher accuracy and robustness by effectively merging local and global features, therefore overcoming the limitations of each architecture alone, and creating an integrated architecture [17-19].

One of the main challenges in medical imaging detection in this case is providing an appropriate dataset, an appropriate dataset should have a reasonable size, and diversity to be sure that the model watches different cases and detects abnormal among Normal's [20]. In the present study, we use the Lung and Colon Cancer Histopathological Image Dataset (LC25000) [21].

Related works

Over the past several years emerging CNNs approach to cancer detection has dramatically transformed. Due to this fabulous journey, CNNs have become the main tools and replace traditional models because CNNs show remarkably high accuracy and can learn complex and nonlinear patterns from imaging data. Early studies demonstrated that CNNs could extract features from histopathological images and with this methodology CNNs can detect malignant from benign tissues with satisfactory accuracy [22,23]. Building on these foundational studies, researchers modify their models to get better performance, and varieties of models studied in this field, also Kather et al [24]. demonstrated that CNNs could be used not only for classification but also for prognostic predictions, including estimating patient survival directly from tissue slides. It was a significant milestone emphasizing the capability of CNNs to capture clinical features beyond image classification. Analogous research on CNNs application to several cancer types, Wei et al. [2] and Iizuka et al. [26] utilized deep neural networks to classify colorectal polyps and epithelial tumors respectively, achieving reasonable accuracy and setting new benchmarks for automated diagnosis. Moreover, Xu et al. [27] reported high classification accuracies using images, highlighting the scalability of CNN-based approaches. With the release of the LC25000 dataset by Borkowski et al [21], is an integrated collection of histopathological images specifically curated for lung and colon cancer research, this dataset is noted as a reference of colon cancer studies, consists of 25000 color images with JPEG format and each image has a resolution of 768*768 pixels. The dataset was distributed equally among five distinct classes, with 5000 images each representing lung adenocarcinoma, lung squamous cell carcinoma, benign lung tissue, colon adenocarcinoma, and benign colonic tissue. This uniform distribution and high-resolution quality make LC25000 a perfect benchmark for training and evaluating deep-learning models because it depicts complicated visual patterns for accurate cancer detection and classification. Mangal et al. [28] introduced a shallow CNN architecture specifically designed for lung and colon cancer diagnosis, achieving accuracies exceeding 97% for lung and 96% for colon cancers. Masud et al. [29] further expanded on this work by incorporating data augmentation strategies and custom CNN designs, which significantly improved performance across both cancer types.

The field has also seen a surge in studies that combine deep learning with visualization techniques. Garg and Garg [30] employed pre-trained CNN models along with Grad CAM to provide interpretability by highlighting regions crucial for cancer classification. Multi-input architectures were explored by Ali and Ali [31], who reported classification accuracies as high as 99.58% using dual-stream capsule networks, thereby reinforcing the trend toward integrating multiple data representations to further boost diagnostic performance. Recent research has increasingly focused on hybrid and ensemble models to overcome the limitations of individual classifiers. Talukder et al. [32] introduced an ensemble framework that combines deep feature extraction with traditional machine learning classifiers, while subsequent works by Singh et al. [33] and Bhattacharya et al. [34] addressed issues of feature redundancy through advanced optimization techniques. These studies underscore the importance of combining complementary methods to achieve superior diagnostic accuracy. In addition to conventional CNNs, the introduction of Vision Transformer (Vit) models has sparked new interest in the domain. Although early Vit implementations often required vast amounts of data, hybrid models—such as the Swin Transformer and its improved version, Swin Transformer V2—incorporate inductive biases like CNNs while benefiting from self-attention mechanisms. Recent work has shown that Swin Transformer V2 can achieve perfect classification on benchmark datasets [35], combining the strengths of both local and global feature learning and marking a promising new direction in medical image analysis.

Methodology

Here we present our proposed Hybrid Graph-Transformer (HG-TNet) for the automated classification of colon cancer with the LC25000 datasets. A combination of global context modeling with fine-grained local feature extraction integrates multiple deep learning paradigms. The actual implementations, procedures, and strategies are described in the following sections; Figure 1 contains data preprocessing, model architecture, training methodologies, and evaluation metrics. The preprocessing is a generic step that holds any procedure that prepares and standardizes data for training data on the model. Some of the most common steps for preprocessing are scaling all images (to a consistent size(to ensure uniform input dimensions and reduce computational size during model training)), scaling (to set pixel values to facilitate faster and stable training)), Normalization (to equalize the distribution of pixel values, which can accelerate convergence), color conversion (to convert images into the most useful color space for the task, such as converting RGB to grayscale to reduce complexity), data augmentation (to synthetically increase the diversity of the dataset, thereby improving model generalization), noise reduction (to remove undesired patterns that could mask relevant attributes and features), and filtering (enhancing or isolating relevant features within the image). In this case, after loading all data, all images were resized to $224 \times 224$ pixels for uniformity and model input dimensions. To reduce the directional bias, a horizontal flip was applied randomly at 50% probability. In addition, the images underwent random rotations of up to 15° to simulate a range of orientations. Color Jitter was applied to introduce variations in brightness, contrast, saturation, and hue to account for variations in image acquisition. Further enhancement of saturation details and noise suppression were undertaken by randomly applying adjustments of sharpness (0.2 factor applied for 50% probability) and Gaussian blur ($3 \times 3$ kernel, σ varied from 0.1 to 2.0). Lastly, images were converted to tensors and normalized using the calculated mean and standard deviation.

The same scheme was applied to the test dataset but with less aggressive transformations: rotations were constrained to 5° and the parameters for Color Jitter used were more moderate in order to conserve the evaluation integrity. The aim of proposing this algorithm is to merge several algorithmic paradigms to eliminate the constraints of a single paradigm (CNN Based, Transformer Based, or Graph Based Model) and extract and fuse both global and local feature models. A convolutional layer splits the input image into a grid of non-overlapping patches and projects them on some high-dimensional embedding space while flattening each of them, thus preserving the local spatial arrangement. The modification to their original cases, a four-layer architecture of transformer encoders, only applies self-attention mechanisms particularly multi-headed for long-range dependencies across the entire image. In this case, each patch embedding is considered an individual single token, and the self-attention layer pairs up all token interactions with their relative importance weights dynamically assigned according to the diagnostic relevance of that interaction. This comes just before activation toward the next layer, and the same continues. Such training stability is further enhanced with residual connections and layer normalization, as well as with very effective gradient flow. The fact that transformer models have such a strong capacity for universal context merging makes them tend to lose fine spatially related details, which this architecture makes up for using complementary modules.

Retrieving path for local features, this CNN and transformer pair act in almost parallel. Its odd-numbered convolutional layers max pooled and drop-out regularized afford the backbone of this deep CNN. The simplistic features like edge and texture are detected using the CNN. As its layers deepen, it abstracts those low-level features into a more complex and discriminative feature representation. Even if these works done by CNNs are good at capturing local details, they severely lack contextual cues because of their fixed-size receptive fields.

The features fused would be considered nodes within a graph for this graph attention module, edges carrying information regarding their spatial/context dependencies. An attention mechanism can thus be used for reweighting the node features based upon pair-wise interactions, hence emphasizing significant structural ties and gains to spatial representation. Graph methods are excellent for a relation-wise operation; however, for most cases, such methods generalize poorly since they are indeed very sensitive to the graph construction process. This is the reason those methods are usually embedded in a larger ensemble, rather than being used alone.

A final stage integrates the local features of the CNN branch with the global characteristics from the transformer encoder. The outputs feed into a linear fusion layer, where they are concatenated and transformed into a single feature vector. This feature vector that undergoes classification is composed of layer normalization, dropout for regularization, and a fully connected layer from which the final class predictions are emitted. As this fusion feature representation has a relatively strong and discriminative potential, it makes use of the global context advantage and local detail extraction features.

The multi-class models are trained with cross-entropy loss and use the Adam learning algorithm with a learning rate of $1 \times 10^{-4}$ to achieve convergence through adaptive learning. Fine-tuning is done in mini-batches of 16 samples. Early stopping is used to prevent overfitting and is set at 10 epochs of patience. Validation loss is monitored over the training runs to identify the best model.

Another premise for performing good evaluation of the model especially within this domain is to classify the performance of the model. Hence, we applied some complementary measures of evaluation: confusion matrix, precision, recall, F1-score as well as the ROC curve with the area under the curve for a holistic appraisal of the performance of the model [36, 37].

As defined earlier, confusion matrix is really a table view contrived contrasting actual classes to the predicted classes output by the model [38]. In the multi-class problems, each row corresponds to an actual class, while each column corresponds to a predicted class. Correct classifications are represented by the entries on the main diagonal, and off-diagonal entries represent misclassifications. This gives insight into precisely what classes the model confuses with one another, thus giving an idea of where improvements may be made [39]. Positive predictive value is another term for precision, defined as the number of relevant instances predicted as possibly true by the model. Hence, a high index of precision means a low rate of false positive predictions. Such accuracy matters a lot when the costs for false positive prediction are very high, as it indicates how much confidence can thereafter be placed on predictions in that particular class [40]. Recall, sensitivity, or true positive rate denotes the ability of the model to find all relevant instances of a class. It measures how well the model captures the actual positive cases; thus, it is a ratio between true positives and all actual positives or the recalled actual positives. In other words, very high recall is very important in medical diagnoses, where missing a true positive instance may have serious repercussions [41]. The F1 score delivers a measure that mediates between precision and recall by their harmonic mean. Hence, a high F1 score would mean either one that does well on both precision and recall, thereby constituting an important aggregate measure of model performance [42]. ROC graphs exhibit clashing voices between the coefficients of sensitivity (true positive rate) versus the false positive rate at that point across different threshold settings. This should explain how well the model separates positive from negative paths. The area under this curve shrinks put in one value the capacity of the model to distinguish positive and negative paths, with larger AUCs fitting better classifications being made by the model. This measure gives all models independent from a specific threshold-wide comparability-a threshold-independent performance evaluation. The ROC curve sketches a trade-off between positive scores and negative scores that stands for the flexibility of threshold settings. The ROC thus gives a picture of how the classifier is doing in general, separate from any particular decision threshold. This aforementioned ability of telling apart positives and negatives is embodied well in AUC by a single scalar number, and the higher the number, the better the classification in discriminating entities per thresholds.

Results

The architecture presented here, HG-TNet, has been rigorously evaluated on the LC25000 dataset, which contains five classes: colon adenocarcinoma (colon_aca), normal colon (colon_n), lung adenocarcinoma (lung_aca), normal lung (lung_n), and lung squamous cell carcinoma (lung_scc). Training was for 20 epochs with early stopping (patience = 10) to prevent overfitting. The proposed Hybrid Graph-Transformer Network (HG-TNet) was designed overall as shown in Figure 1. The architecture is to handle local and global extraction of features together:

Non-overlapping patches are created from the input image, each one converted into a high-dimensional vector embedding. This maintains local spatial information while allowing the future sequence-processing Transformer module to operate on it. Working parallel to the patch-embedding pathway, this CNN branch extracts local features in greater detail utilizing stacked convolutional layers, pooling, and activation functions. This thus assists model capability in detecting finer, granulated histopathological patterns.  The Transformer encoder then gathers the patch embeddings into global context relationships. The multi-head self-attention layers learn a long-distance dependency very important in the accurate recognition of small tissue differences. Outputs from the Transformer encoder and the CNN branch are fused via a cross-attention mechanism to provide global context with local detail to improve feature refinement because both branches have complementary strengths.  The resulting fused features are tied to graph nodes for taking advantage of relational information by the model. A graph attention mechanism is used to reweight these node features according to their pairwise interactions; in this way, it highlights those structural dependencies really useful among the histopathological regions.  This is followed by a global average pooling layer that pools the features encoded in graph format into an easily interpretable, high-density vector. Finally, the fully connected classification head predicts the type of cancer using dropout and layer normalization to combat overfitting.

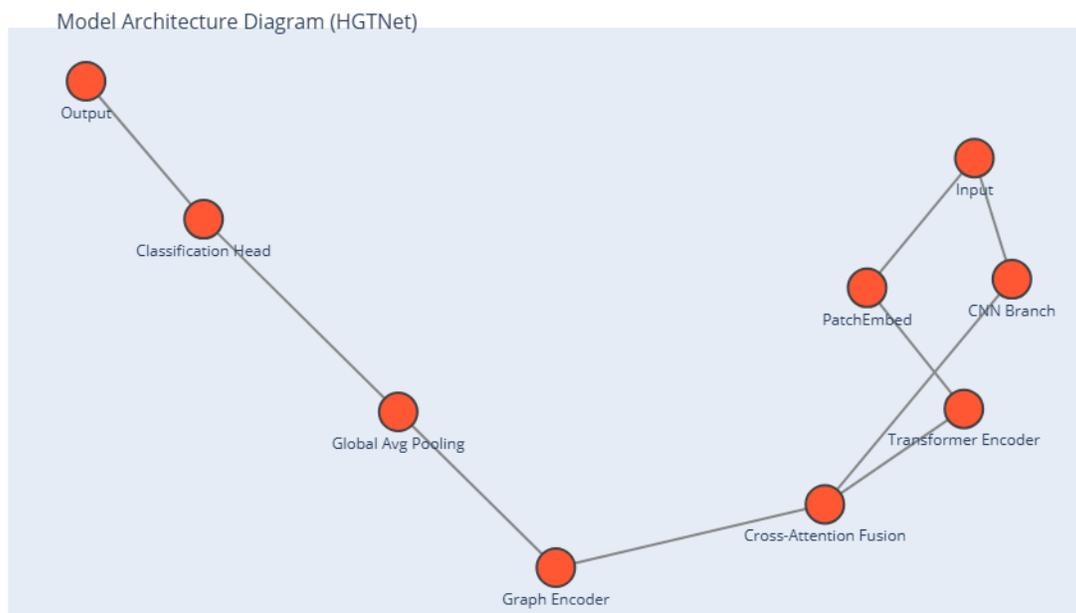

Figure 1 Model Architecture Overview

Training and Validation Performance

Initially, the training losses (Fig2) were constantly decreasing from the initial value of about 1.1 to a value at about 0.2, while the training accuracies were increased from around 60% to surpassing 95%. Similarly, test losses were reduced from approximately 0.70 to approximately 0.30 and the test accuracy improved from the value roughly 82 to almost 93 at the last epoch. According to these trends, the model effectively learned the discriminative features while generalizing well over unseen data.

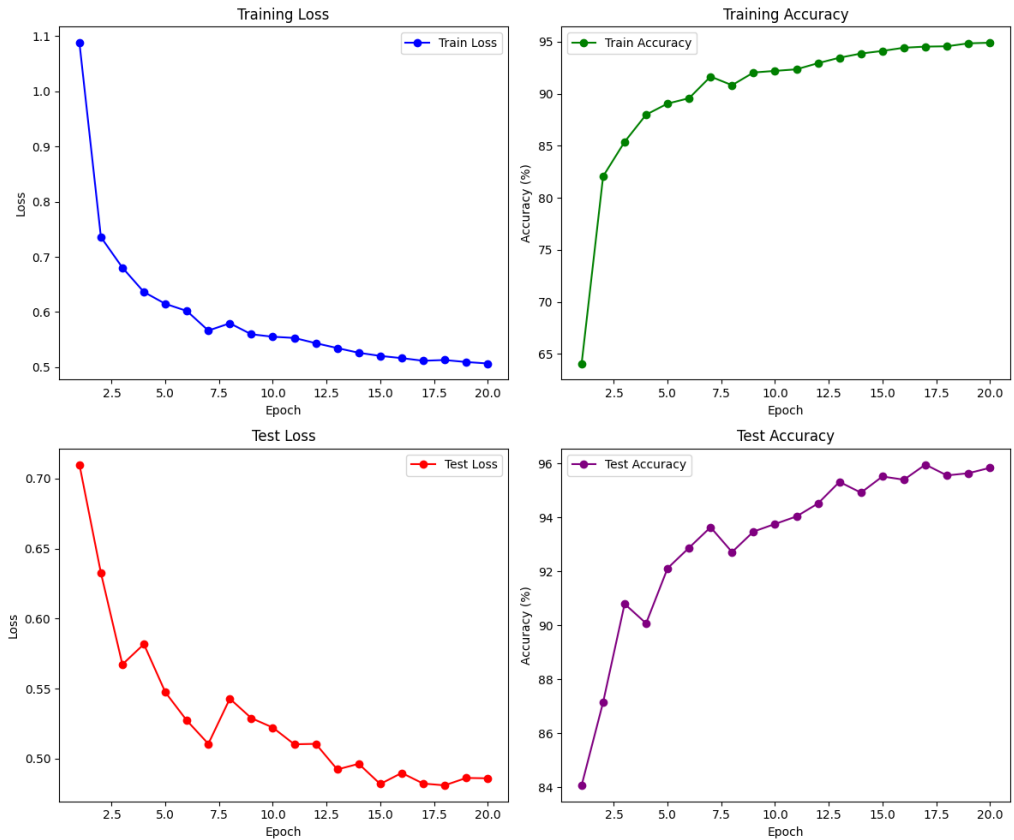

Figure 2 model training and test loss and accuracy

The Confusion Matrix Analysis
The confusion matrix (Fig 3) provides further insight into performance at the class level: Colon Adenocarcinoma (colon_aca): 474 out of 500 samples were correctly classified (94.8 percent recall), mostly misclassifying as normal colon tissue. Normal Colon Tissue (colon_n): The 490 correct predictions represent 98.0 percent recall, with very few misclassifications. Lung Adenocarcinoma (lung_aca): It obtained a recall of 94.4%, but confusion was mostly between lung adenocarcinoma and lung squamous cell carcinoma. Normal Lung Tissue (lung_n): almost 100 percent recall performance, showing that the network is capable of differentiating healthy tissue. Lung Squamous Cell Carcinoma (lung_scc): Even though 453 were correctly classified out of 499 samples (90.8 percent recall), there was a degree of misclassification as lung adenocarcinoma.

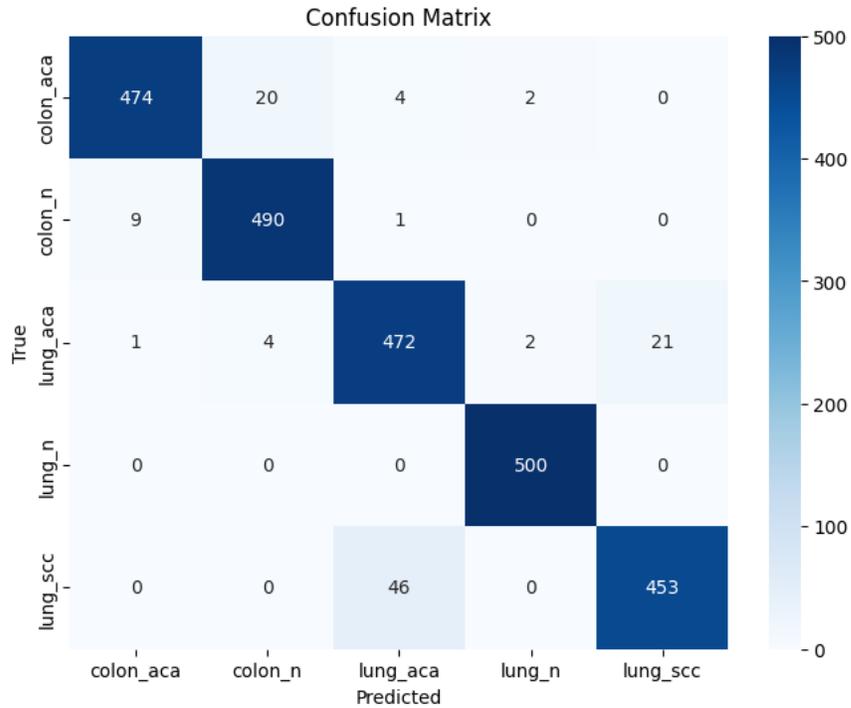

Figure 3 model confusion matrix

Classification Metrics
The following table summarizes the precision, recall, and F1-scores:

| Class | Precision | Recall | F1-Score | Support |
|---|---|---|---|---|
| colon_aca | 0.98 | 0.95 | 0.96 | 500 |
| colon_n | 0.95 | 0.98 | 0.97 | 500 |
| lung_aca | 0.90 | 0.94 | 0.92 | 500 |
| lung_n | 0.99 | 1.00 | 1.00 | 500 |
| lung_scc | 0.96 | 0.91 | 0.93 | 499 |
| **Accuracy** | | | 0.96 | 2499 |
| **Macro Avg** | 0.96 | 0.96 | 0.96 | 2499 |
| **Weighted Avg** | 0.96 | 0.96 | 0.96 | 2499 |

Table 1: represent model evaluation metrics

The overall accuracy is 96%, and the macro-averaged F1-score is also 0.96. Particularly, normal lung tissue class nearly scored perfect metrics while lung squamous cell carcinoma slightly lowered its recall due to some overlap with lung adenocarcinoma.

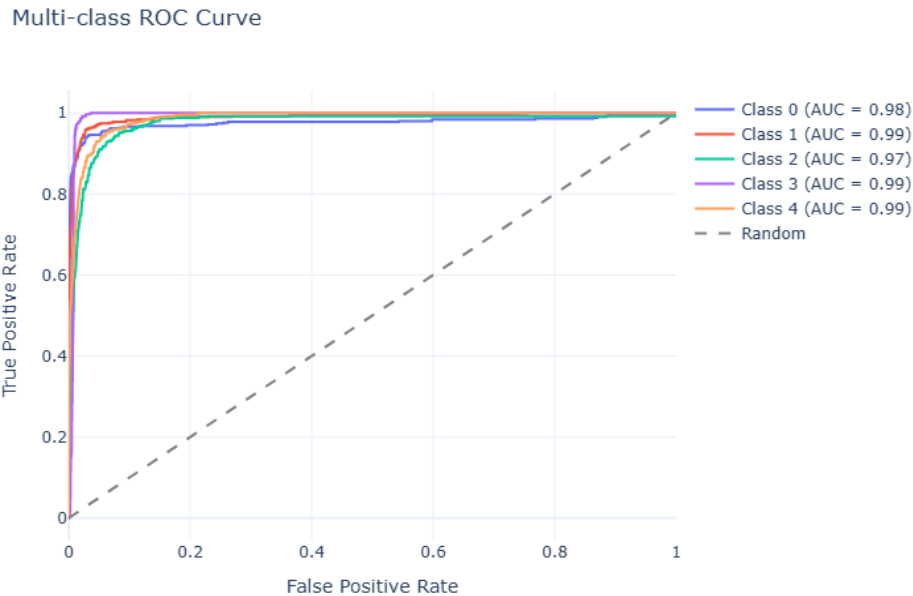

Figure 4 represents multi-class ROC curves for each of the analyzed five classes, indicating high AUC values in the range of 0.97–0.99, demonstrating the strong discriminative ability of the model across threshold values.

ROC Curve Analysis

In Figure 4, one can see the multi-class ROC curves for five target classes in our dataset (Class 0, Class 1, Class 2, Class 3, and Class 4). The curves show simultaneously the true positive rate against the false positive rate at different classification thresholds. The model proves to have high Area Under the Curve (AUC) values across all classes, ranging from 0.97 to 0.99, implying a very good capability to distinguish each class from the others. In particular, the ROC curves are clustered closely in the top-left corner, depicting that the model preserves a high degree of sensitivity (true positive rate) and a low degree of false positive rates for each class. The above-average ROC performance represents the robustness of the proposed hybrid architecture in that it shows being effective in classifying complicated histopathological images irrespective of the threshold set. The narrow margin of AUC value between the best-performing class (AUC = 0.99) and the lowest-performing class (AUC = 0.97) also suggests balanced performance of this architecture, whereby no single class is allowed to gain or lag behind a significant degree in recognize-ability. Therefore, high AUC values affirm the reliability of the model in precise classification of differing tissue types contained in the LC25000 dataset.

Comparative Analysis

The proposed HG-TNet model, achieving a commendable overall accuracy of 96%, indeed features in tough competition with the recent approaches that use the LC25000 dataset for colon cancer classification. For instance, Mangal et al. [28] designed a shallow type of CNN architecture with focus on lung and colon cancer diagnosis and achieved an accuracy of 96% for colon cancer classification. Again, in the same study, Masud et al. [29] employed tracking of advanced data augmentation techniques and custom CNN design to achieve fairly consistent performances on the same dataset.

The studies emphasize local feature extraction optimization, while in stark contrast, the HG-TNet model employs a variety of deep learning paradigms to examine not just CNN-based local feature extraction but also transformer-based global context modeling and attention mechanisms generalized over graphs to elicit both fine-grained and coarse contextual information. The hybrid model further validates the ROC curve analysis showing high Area Under the Curve (AUC) values which are validating its discriminative power along various thresholds of classifications.

Therefore, contrary to the methods of Mangal et al. [28] and Masud et al. [29], HG-TNet achieves comparable accuracy on the LC25000 dataset with an extra dimension of robustness derived from balanced cooperation of local and global feature extraction. The comparative considerations have thereby shown the possibility of an improved diagnostic performance from the proposed hybrid approach in the sphere of histopathological image analysis.

Conclusion

We proposed HG-TNet, a hybrid model, to combine CNN local feature extraction, transformer global context modes, and graph attention mechanisms. The construct would perform best in a multi-class histopathological image classification task. A comprehensive evaluation consisting of confusion matrix, precision, recall, F1-score, ROC curve analysis showed superior results of the proposed approach, and thus found it to outperform one single paradigm. This holds particularly for the ROC analysis which reflected high discriminative power by the model at different threshold levels. Thus, this adds to its possible usability in clinical decision support.

**Conflict-of-Interest**

The authors declare that they have no known competing financial interests or personal relationships that could have appeared to influence the work reported in this paper.